\documentclass{article}

 \PassOptionsToPackage{numbers, compress}{natbib}
\usepackage[final]{nips_2016}
\usepackage[utf8]{inputenc} 
\usepackage[T1]{fontenc}    
\usepackage{hyperref}       
\usepackage{url}            
\usepackage{booktabs}       
\usepackage{amsfonts}       
\usepackage{nicefrac}       
\usepackage{microtype}      
\usepackage{url}
\usepackage{graphicx}

\title{Skin Lesion Classification Using Deep Multi-scale Convolutional Neural Networks}

\author{
  Terrance DeVries \\
  School of Engineering,\\
  University Of Guelph\\
  Guelph, N1G 2W1 ON, Canada \\
  \texttt{terrance@uoguelph.ca} \\
   \And
  Dhanesh Ramachandram \\
  School of Engineering,\\
  University Of Guelph\\
  Guelph, N1G 2W1 ON, Canada \\
  \texttt{dramacha@uoguelph.ca} \\
}

\begin{document}

\maketitle

\section{Executive Summary}
 We present a deep learning approach to the ISIC 2017 Skin Lesion Classification Challenge 
 using a multi-scale convolutional neural network. Our approach utilizes
 an Inception-v3 network pre-trained on the ImageNet dataset, which is 
 fine-tuned for skin lesion classification using
 two different scales of input images. 

\section{Background}

 Melanoma is a malignant tumour originating from melanocytes cells - skin cells 
 responsible for the production of melanin. The American Cancer Society 
 estimates that in the United States alone for 2017, more than 
87,000 new melanoma cases will be diagnosed and around 9,300 persons are 
expected to die\cite{acs2017}.  Skin melanoma lesions are very challenging to 
visually diagnose due to their similarity in visual characteristics with other 
benign skin lesions such as nevus and seborrhoeic keratosis.  

 Expert dermatologists review dermoscopy images using skin surface microscopy 
(also called `epiluminoscopy' or `epiluminescent microscopy'). 
 Dermoscopy is mainly used to evaluate pigmented lesions in order to 
 distinguish malignant skin lesions, such as melanoma and pigmented basal cell 
 carcinoma, from benign melanocytic naevi and seborrhoeic keratoses.  The most 
 common imaging technique is the fluid immersion method, which involves applying 
 mineral oil or spraying alcohol onto the lesion and then placing the lens, 
 typically with 10x- 14x magnification, in contact with the skin. 
 
Dermatologist have devised several techniques to classify lesions. The 3-point 
checklist flags a lesion as melanoma if two out of three following features are 
found on lesions\cite{soyer2004three}:
\begin{itemize}
\item \textbf{Asymmetry}: asymmetry of color and structure in one or two perpendicular 
axes
\item \textbf{Atypical network}: pigment network with irregular holes and thick lines

\item \textbf{Blue-white structure}s: any type of blue and/or white colour, i.e. 
combination of blue-white veil and regression structures
\end{itemize}
Another technique used by dermatologists for screening for melanomas is known 
as the \textbf{ABCD} parameters\citep{johr2002dermoscopy}:
\begin{itemize}
	\item \textbf{Asymmetrical shape} - melanoma lesions are typically asymmetrical
	\item \textbf{Borders} - melanoma lesions have irregular border
	\item \textbf{Color} - presence of more than one color in melanoma lesions
	\item \textbf{Diameter} - melanoma lesions are typically larger than 6mm in diameter	
\end{itemize}

In recent years, there have been a number of computerized diagnosis techniques developed which use digital image processing in an attempt to classify benign and malignant skin lesions from dermoscopy 
images\citep{celebi2015dermoscopy}. To accelerate the innovation in this area the International Skin Imaging Collaboration (ISIC)
\footnote{\url{http://isdis.net/isic-project/}}
has initiated the ISIC Challenge. This challenge provides  
a set of 2,000 publicly available dermoscopic images to participants to apply 
computer vision and machine learning techniques to three different tasks: lesion segmentation, dermoscopic feature extraction, and lesion classification. Our approach aims to address the challenge of lesion classification.

\begin{figure}
\includegraphics[width=\textwidth]{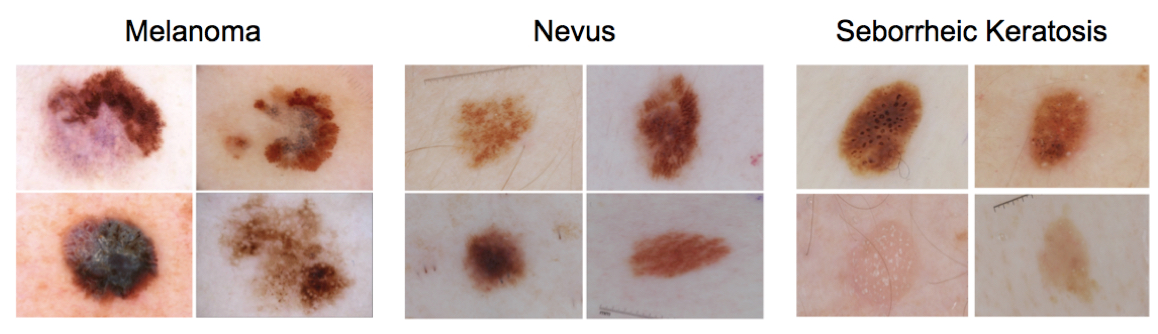}
\caption{Examples of skin lesions from the ISIC 2017 Skin Lesion Classification Challenge dataset}
\label{example_skin_lesions}
\end{figure}

The lesion classification portion of the challenge is composed of two independent binary image classification tasks that involve three unique diagnoses of skin lesions (melanoma, nevus, and seborrheic keratosis). The first binary classification task involves distinguishing  (a) melanoma vs. (b) nevus and seborrheic keratosis. In the second binary classification task, the classifier needs to distinguish between (a) seborrheic keratosis and (b) nevus and melanoma.
\textit{Melanoma}, as described earlier, is a malignant skin tumor, derived from melanocytes (melanocytic), \textit{Nevus} is benign skin tumor, derived from melanocytes (melanocytic) while \textit{Seborrheic keratosis} is also a benign skin tumor, derived from keratinocytes (non-melanocytic).
Examples from the dataset of these three skin lesions are shown in Figure \ref{example_skin_lesions}.

Deep neural network models, particularly deep convolutional neural networks and its variants, are currently
the best performing image classification models for a wide variety of tasks and applications. They have made significant impact in multiple problem domains,
including computer-aided diagnosis using medical images. A recent paper published in Nature \cite{esteva2017} is the testament to the success
of deep convolutional neural networks for detecting melanoma from dermoscopy images, achieving an AUC of 0.94 on this task. A critical factor that contributed to the high level of performance of that particular model was the availability of a large dataset, containing more than 120,000 images (60x larger than available for the ISIC Challenge).  For this challenge, our deep-learning model is similar to what was used by \cite{esteva2017}, however since we only have access to a much smaller dataset we apply a number of optimization and post-processing techniques to improve the performance of our classifier.

\section{Methodology}

In this extended abstract, we describe the architectural details of the deep multi-scale neural network model used for our submission to this challenge. Our methodology involves fine-tuning an ImageNet pre-trained Inception-v3 \cite{inceptionv3} deep neural network on the ISIC 2017 dataset using two input images at different scales of resolution. The two scales correspond to a coarse scale that captures the overall context and shape characteristics of the lesion, while the image at the finer scale reveals textural details and various low-level characteristics of the lesion that are important for distinguishing between the classes of lesion.

\begin{figure}
\includegraphics[width=\textwidth]{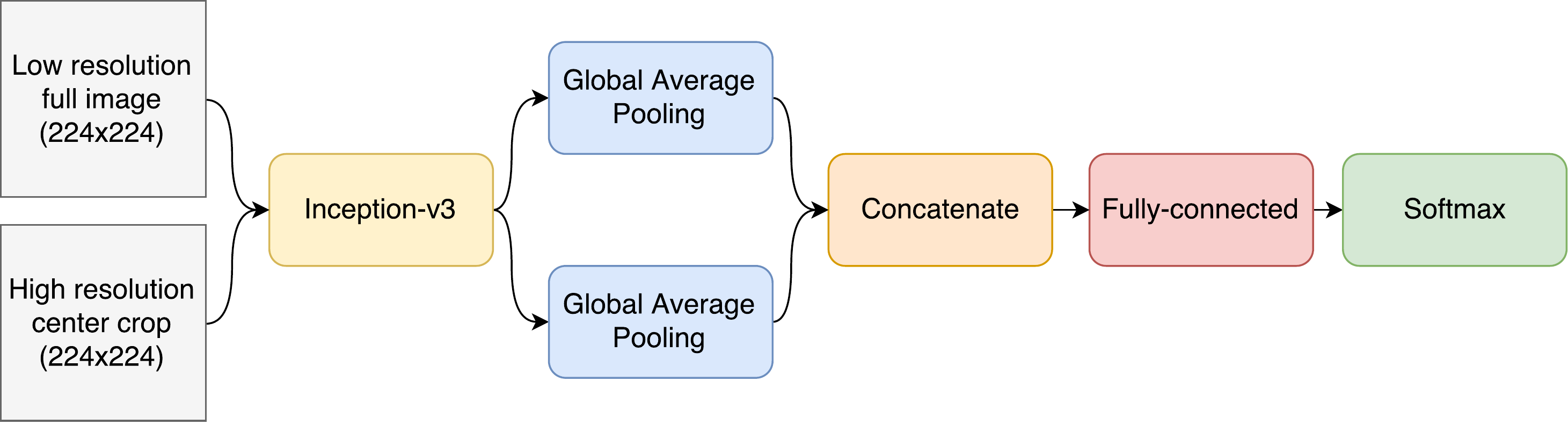}
\caption{Multi-scale CNN model architecture}
\label{model_architecture}
\end{figure}

\subsection{Data Preprocessing}
The multiscale network receives input images at two scales. At the coarse scale the input images are simply resized to $224\times224$, while at the finer scale input images are first resized to $448\times448$ and then a center crop of size $224\times224$ is extracted. Pixel values are re-scaled to a range of $0-1$. No further pre-processing is applied to the images prior to training or inference.

\subsection{Additional Data}

We found it beneficial to augment our dataset with additional training data from the ISIC\_MSK\-2\_1 dataset\footnote{\url{https://isic-archive.com/\#images}}, which contained 334 melanoma images, 3 seborrheic keratosis images, and 1004 nevus images, in order to improve the results of melanoma prediction. As this dataset included very few extra examples of seborrheic keratosis, we found that it did not help improve performance for that class.

\subsection{Architecture}

For the base of our network we used an Inception-v3 model pre-trained on ImageNet. All fully-connected layers from the Inception network were removed so that only the convolutional feature extraction layers remained. After the convolutional layers we added a global average pooling layer to condense the output feature maps into a 2048 element feature vector. 

The multi-scale network is created by first passing both the low resolution image and high resolution center-crop image through the same Inception feature extractor. The resulting feature vectors from each image are concatenated to produce a single 4096 element vector, which is then passed through a fully-connected layer with 1024 hidden units and ReLU activations. Finally, a 3-way softmax is used to generate probability predictions corresponding to each of the three classes: melanoma, seborrheic keratosis, and nevus. The full architecture is demonstrated in Figure \ref{model_architecture}. Implementing the network so that inputs share a single Inception feature extraction branch encourages the model to learn features that are useful at multiple scales, and is much more memory efficient than using a separate Inception branch for each different resolution input.

\subsection{Training}
During training we initially freeze all layers in the base Inception network and only apply weight updates to the newly added fully-connected layers. This step is crucial before the fine-tuning stage so that the pre-trained weights in the convolutional layers are not immediately destroyed by large gradients from the untrained fully-connected layers. For this step we use the Adam optimizer \citep{kingma2014adam} with an learning rate of $10^{-2}$ and perform 150 weight updates.

For the fine-tuning stage, the learning rate is decreased to $10^{-3}$ and the last two inception modules of the Inception network are also unfrozen. The weights for all other layers remain frozen during training. While unfreezing all layers of the Inception network also yielded reasonable results, we decided against this in an attempt to prevent overfitting due to the small dataset size. We perform 3500 weight updates in the fine-tuning stage. 

In both stages of training, we use mini-batches of 32 images, which are generated by randomly selecting an equal number of samples from each class by oversampling the minority classes. Both techniques, using a small batch size and randomly sampling data (with replacement) instead of iterating through the entire dataset, appeared to combat overfitting on the training set. 

Further regularization is applied by performing data augmentation during training. We found that randomly applying rotations of 90 degree increments and randomly mirroring the images improved performance of the model. Experiments were conducted with other forms of data augmentation, such as rotating images randomly from 0 to 360 degrees or applying random shifting, but we did not observe any benefit from these transformations.

\subsection{Post-processing}
During inference, for each test image, we apply 90 degree rotations and mirroring to get 8 augmented images which are then passed each through the trained network. The final prediction probability is the averaged prediction from these 8 augmented images. As the challenge is binary classification, we convert the prediction from 3-class to binary classification by a one-vs-rest approach.

\subsection{Ensembling}

To further improve the performance of our final model, we trained several models with slight variations on the original formulation in order to increase the variance of the outputs. The final predictions from each model were then combined using geometric averaging. In total, we combined the outputs from 10 different models. Variations applied to the previously described model include increasing the batch size, increasing the number of inception modules that were unfrozen, and training on different folds of the dataset. A single-scale model trained on images resized to 448x448 instead of 224x224 was also added to the ensemble. 

\section{Results}
Table \ref{table:results} shows the performance of our model when evaluated on the official ISIC 2017 Skin Classification Challenge validation set. 

\begin{table}[h]
\centering
\caption{Accuracy and AUC for full ensemble evaluated on official validation set.}
\label{table:results}
\begin{tabular}{|l|c|c|c|}
\hline
         & \textbf{Melanoma} & \textbf{Seborrheic Keratosis} & \textbf{Average} \\ \hline
\textbf{Accuracy} & 0.893    & 0.913                & 0.903   \\ \hline
\textbf{AUC}      & 0.896    & 0.990                & 0.943   \\ \hline
\end{tabular}
\end{table}

\bibliographystyle{unsrt}
\bibliography{references}

\begin{thebibliography}{1}

\bibitem{acs2017}
American~Cancer Society.
\newblock Key statistics for melanoma skin cancer, 2017.
\newblock Accessed: 25 Feb 2017.

\bibitem{soyer2004three}
H~Peter Soyer, Giuseppe Argenziano, Iris Zalaudek, Rosamaria Corona, Francesco
  Sera, Renato Talamini, Filomena Barbato, Adone Baroni, Lorenza Cicale,
  Alessandro Di~Stefani, et~al.
\newblock Three-point checklist of dermoscopy.
\newblock {\em Dermatology}, 208(1):27--31, 2004.

\bibitem{johr2002dermoscopy}
Robert~H Johr.
\newblock Dermoscopy: alternative melanocytic algorithms—the abcd rule of
  dermatoscopy, menzies scoring method, and 7-point checklist.
\newblock {\em Clinics in dermatology}, 20(3):240--247, 2002.

\bibitem{celebi2015dermoscopy}
M~Emre Celebi, Teresa Mendonca, and Jorge~S Marques.
\newblock {\em Dermoscopy image analysis}.
\newblock CRC Press, 2015.

\bibitem{esteva2017}
Andre Esteva, Brett Kuprel, Roberto~A Novoa, Justin Ko, Susan~M Swetter,
  Helen~M Blau, and Sebastian Thrun.
\newblock Dermatologist-level classification of skin cancer with deep neural
  networks.
\newblock {\em Nature}, 542(7639):115--118, 2017.

\bibitem{inceptionv3}
Christian Szegedy, Vincent Vanhoucke, Sergey Ioffe, Jonathon Shlens, and
  Zbigniew Wojna.
\newblock Rethinking the inception architecture for computer vision.
\newblock {\em CoRR}, abs/1512.00567, 2015.

\bibitem{kingma2014adam}
Diederik Kingma and Jimmy Ba.
\newblock Adam: A method for stochastic optimization.
\newblock In {\em The International Conference on Learning Representations
  (ICLR)}, 2015.

\end{thebibliography}

\end{document}